# FedGlu: A personalized federated learning-based glucose forecasting algorithm for improved performance in glycemic excursion regions

Darpit, Dave

Department of Industrial and Systems Engineering, Texas A&M University, darpitdave@tamu.edu

Kathan Vyas

Department of Computer Science and Engineering, Texas A&M University, darpitdave@tamu.edu

Jagadish Kumaran Jayagopal

Department of Computer Science and Engineering, Texas A&M University, darpitdave@tamu.edu

Alfredo Garcia

Department of Industrial and Systems Engineering, Texas A&M University, darpitdave@tamu.edu

Madhav Erraguntla

Department of Industrial and Systems Engineering, Texas A&M University, darpitdave@tamu.edu

Mark Lawley*

Department of Industrial and Systems Engineering, Texas A&M University, darpitdave@tamu.edu

Continuous glucose monitoring (CGM) devices provide real-time glucose monitoring and timely alerts for glycemic excursions, improving glycemic control among patients with diabetes. However, identifying rare events like hypoglycemia and hyperglycemia remain challenging due to their infrequency. Moreover, limited access to sensitive patient data hampers the development of robust machine learning models. Our objective is to accurately predict glycemic excursions while addressing data privacy concerns. To tackle excursion prediction, we propose a novel Hypo-Hyper (HH) loss function, which significantly improves performance in the glycemic excursion regions. The HH loss function demonstrates a 46% improvement over mean-squared error (MSE) loss across 125 patients. To address privacy concerns, we propose FedGlu, a machine learning model trained in a federated learning (FL) framework. FL allows collaborative learning without sharing sensitive data by training models locally and sharing only model parameters across other patients. FedGlu achieves a 35% superior glycemic excursion detection rate compared to local models. This improvement translates to enhanced performance in predicting both, hypoglycemia and hyperglycemia, for 105 out of 125 patients. These results underscore the effectiveness of the proposed HH loss function in augmenting the predictive capabilities of glucose predictions. Moreover, implementing models within a federated learning framework not only ensures better predictive capabilities but also safeguards sensitive data concurrently.

---

# 1 INTRODUCTION

A peptide hormone known as insulin is produced by the beta cells in the pancreases. Insulin plays a crucial role in regulating the transport of glucose and facilitating its absorption by cells from the bloodstream for energy production. Any disruption in this process leads to a medical condition called Diabetes mellitus (DM), more commonly referred as diabetes. Globally, this chronic disease affects nearly 537 million lives with an additional 374 million classified as prediabetic, putting them at similar risks of developing complications [1]. Diabetes can lead to both, immediate and long-term health issues, including blindness, kidney failures, amputations, heart-related diseases, seizures and in severe cases, even death [2-4]. Diabetes is broadly categorized into Type 1 and Type 2. In type 2 diabetes, cells become resistant to insulin, hindering their ability to absorb glucose. Type 2 diabetes constitutes 95% of the diabetes population and is typically prevalent in older age-groups. On the other hand, Type 1 diabetes arises when the pancreas fails to produce sufficient insulin. Type 1 diabetes accounts for about 5% of the diabetes population and is mainly found in age groups 0-20 years of age. However, it requires a much more rigorous diabetes management to avoid the imminent consequences. The overarching goal of diabetes management is to maintain euglycemia (normal glucose levels) or increase time spent within the optimal range (time-in-range, TIR)[2]. Insulin therapy is the mainstay of diabetes management, requiring a delicate balance between short-term and long-term objectives [5, 6]. While diabetes cannot be cured, proper insulin management allows patients with diabetes to keep optimal blood glucose levels and minimize the risk of further complications [7].

The most widely used method for glucose monitoring involves the use of fingerstick technique, where a small blood sample is drawn by pricking the finger and analyzed using the glucometer [8, 9]. This was revolutionized by Continuous Glucose Monitoring (CGM) devices, that offer real-time glucose monitoring with automated readings. This has evidently led to increasing time spent by patients in TIR, a crucial metric in diabetes care [10, 11]. CGM devices have proven to enhance glycemic control by reducing both hypoglycemia and hyperglycemic excursions through real-time predictions [12, 13]. While previous research for Type 1 diabetes has primarily focused on predicting hypoglycemic events [10, 14-19], achieving glycemic goals requires maximizing glucose readings in TIR by avoiding both hypoglycemia and hyperglycemia.

Efforts to develop machine learning (ML) models for glucose predicting using historical CGM readings have shown reasonable efficacy. These models typically rely on extensive training data, often gathered through large-scale clinical research studies or healthcare providers. The data is often stored in cloud-based servers and poses significant privacy risks, with concerns about data theft and user privacy [20]. Given the escalating concerns of data theft, user security and privacy, recently many countries have enforced regulations aiming to address these concerns and protection regulations to protect the interests of the people [21-23]. An alternative strategy is to develop individual models with personal data to enhance privacy, but this comes with the drawback of reduced prediction performance due to limited data. ML and deep learning models require substantial data for optimal performance.

Thus, to overcome the challenges in glycemic excursion prediction and privacy concerns, we propose a collaborative learning framework that addresses these concerns simultaneously. The major contributions of this paper are as follows:

- A novel, HH loss function, for improved prediction performance in the glycemic excursion regions
- FedGlu, a machine learning model trained in a federated learning framework for improving model performance and data privacy preservation simultaneously.

The remainder of the paper is organized as follows: Section II reviews the current literature and points out relevant papers. Section III describes the dataset in detail. It also explains the methods like HH loss function, and the federated learning used as part of the paper, different metrics used for evaluation and the different setup of experiments in the paper. Section IV compares the results derived for the different experiments with the HH loss function, comparison between central, local, and personalized federated models. Section V provides discussion and insights on the results, limitations, and future work for the paper. Lastly, Section VI concludes the paper.

## 2 STATE OF THE ART

In this section, we provide a brief overview of the existing literature on glucose prediction. While a comprehensive review may be out of the scope of this paper, we emphasize works that have focused on predicting glucose values with deep learning approaches, use of federated learning for healthcare applications and extending the global federated model through fine tuning for personalization.

### 2.1 CGM-based glucose prediction

The first attempt to predict future glucose levels using past values dates back to 1999 by Bremer and Gough [24]. Since then, researchers have continually advanced the literature to develop powerful and highly accurate models. Machine learning models for glucose prediction fall into two categories: (a) Classification tasks, (b) Regression task. Consistent with the theme of our paper, we focus on the regression tasks. Earlier approaches utilized conventional machine learning methods such as linear regression (LR), support vector regression (SVR), random forests (RF), boosting algorithms [25-28] and time-series forecasting methods like autoregressive integrated moving average (ARIMA) [29, 30] for predicting glucose levels. In recent years, there has been a shift towards employing deep learning (DL) models, leveraging automated feature learning, robust pattern recognition, and abstraction through multiple layers.

Various DL architectures, including multi-layer perceptron (MLP) [31, 32], convolutional neural networks (CNN) and convolutional recurrent neural networks (CRNN) [33, 34], recurrent neural networks (RNN) [35, 36], short long-term memory networks (LSTM) [37-39] , dilated RNNs [40, 41] and bi-directional LSTMs [42-44], have been proposed for blood glucose prediction.

Shuvo et al. [45] introduced a deep multi-task learning approach using stacked LSTMs to predict personalized blood glucose concentration. The proposed approach includes a combination of stacked LSTMs to learn generalized features across patients, clustered hidden layers for phenotypical variability in the data and subject-specific hidden layers for optimally fine-tuning models for individual patient improvement. The authors demonstrate superior results compared to state-of-the-art ML and DL approaches on the OhioT1DM dataset.

A transformer based on an attention mechanism was recently proposed to forecast glucose levels and hypoglycemia and hyperglycemia events [46]. The proposed transformer network includes an encoder network to perform the regression and classification tasks under a unified framework, and a data augmentation step using a generative adversarial network (GAN) to compensate for the rare events of hypoglycemia and hyperglycemia. Results were demonstrated on two datasets, one including Type 1 diabetes patients and the other on Type 2 diabetes patients.

In pursuit of enhanced prediction performance, researchers have often grappled with the constraint of limited data for individual subjects or patients. A recently proposed approach addresses this challenge through multitask learning for advancing personalized blood glucose prediction [47]. This approach was evaluated against sequential transfer learning, revealing two key findings: (a) individual patient data alone may not suffice for training DL models and (b) a thoughtful strategy is crucial to leverage population data for improved individual models. The dataset that was used in this study was the OhioT1DM dataset.

While the literature commonly employs standard regression metrics like root-mean squared error (RMSE), mean absolute error (MAE), and Clark's Error Grid Analysis (EGA) for clinical context, an often-overlooked aspect in the analysis of these metrics in glycemic excursion regions. For instance, RMSE in hypoglycemia and hyperglycemia ranges can be evaluated, Clark's EGA can be applied to glucose readings falling in Zones C,D, and E. Mu et al. [48] introduced a normalized mean-squared error (NMSE) loss function, demonstrating a substantial reduction in RMSE for the hypoglycemia range. However, comprehensive details regarding its superiority over MSE in detecting hypoglycemia are lacking, and its performance in hyperglycemic ranges is not addressed.

### 2.2 Data imbalance

There is a significant imbalance in glucose data distribution across the hypoglycemic, hyperglycemia, and normal ranges. Typically, only 2-10% of glucose readings fall into the hypoglycemic range, while about 30-40% fall in the hyperglycemic range. From a statistical perspective, this presents a classic case of an imbalanced regression problem. While numerous approaches have been developed to tackle the imbalanced data in the classification setting, very few works have been proposed to address the imbalance regression problem, like ours, in the literature [49]. Existing approaches for imbalanced regression can be broadly categorized into two types:

*2.2.1 Sampling-based approaches*

These methods attempt to either undersample the high-frequency values or oversample the low frequency (rare) values. However, determining the 'notion or rarity' in a regression problem is challenging compared to a classification task. Oversampling may lead to overfitting, whereas under sampling may result in sub-optimal performance because of loss of key information. Chawla et. al [50] proposed an approach that generates synthetic samples by combining oversampling and under sampling of the training data for classification task.

*2.2.2 Cost-sensitive approaches*

Cost-sensitive approaches: These approaches introduce a penalizing scheme during training to enable the model to handle outlier values (low-frequency or rare values) enhancing its effectiveness in predicting within those ranges. The recent success of this approach [51-53] motivates us to explore this approach further through our customized loss function.

### 2.3 Federated learning for healthcare

Federated learning (FL) has substantial disruptive potential in healthcare, a domain constrained by sensitive data and strict regulations such as the Health Insurance Portability and Accountability Act (HIPAA) in the US. The reluctance of healthcare entities to share sensitive data has fueled the adoption of FL in various healthcare applications, like medical image processing [54-56], IoT-based smart healthcare applications [57], managing electronic health records (EHR) [58], disease prediction [59-61], predict hospitalizations and mortality [62-64], natural language processing from clinical notes [65-67], etc. A few researchers comprehensively reviewed federated learning in healthcare [68-71].

In the diabetes literature, FL has recently gained traction. A recent study proposed a deep learning approach in the Diabetes Management Control System (DMCS) [72], using 30 virtual subjects from the FDA approved UVA/Padova Type 1 diabetes simulator. Features such as past glucose values, carbohydrate intake, and insulin-on-board were used as input for the model for the diagnosis of diabetes. The findings clearly demonstrate the superior performance of the federated model over the local models.

Another study employed an FL-inspired Evolutionary Algorithm (EA) for classifying glucose values into different risk categories [73] using data from 12 patients in the OhioT1DM dataset. The input features include previous CGM readings, carbohydrates, and insulin data. The results indicate improved performance over local models using an FL-based EA. However, the study's limitation lies in its small sample size of only 12-patients and the absence of addressing real-time glucose-related risk-prediction and its clinical significance.

A decentralized privacy-protected federated learning approach was applied to predict diabetes-related complications using patient-related comorbid features extracted from International Classification of Disease (ICD) codes in real-world clinical datasets [74]. For this, a logistic regression model, 2-layer multi-perceptron model and 3-layer multi-perceptron models were proposed and compared against a centralized (population-level) model. The study addressed class imbalance using techniques like under-sampling, oversampling, and balancing. The results indicate that models developed through the federated learning framework can achieve promising performance that is comparable to the centralized models.

## 2.4 Personalized federated learning

Federated learning has its many unique challenges [75-77], with one prominent issue being the variation in data distribution across clients in the network. This is characteristic of *non-i.i.d.* and imbalanced data [78]. This is particularly significant in healthcare applications where individual patients possess diverse demographics and health histories, necessitating adaptive solutions tailored to each participant [79].

One of the pioneering works in federated learning with wearable healthcare data is *FedHealth* [80], which introduces personalization through transfer learning. This is achieved by first training a conventional global model in a federated learning framework and later fine-tuning two fully connected layers (through transfer learning) to learn activities and tasks for specific users. The study utilizes publicly available human activity recognition data from accelerometry and gyroscope data across 30 users for multiple activity class prediction. The authors employ a CNN based deep learning model, comparing its performance against traditional machine learning methods like RF, SVM, and KNN. The results demonstrate a 4% average improvement in performance for personalized models compared to the global model.

Another noteworthy application of personalized federated learning was used for in-home health monitoring [81]. The authors introduce FedHome, a cloud-edge based federated learning framework, where a shared global model is initially learned from multiple network participants. Individual personalization is achieved using a generative convolutional autoencoder (GCAE), aiming to generate a class-balanced dataset tailored to individual client's data. FedHome exhibits a notable improvement of over 10% in accuracy compared to a conventional global federated learning model.

A recent study on remote patient monitoring (RPM) introduced FedStack architecture, a personalized federated learning approach [79]. The study was based on the MHEALTH dataset with 10 patients [82, 83]. A total of 21 features are extracted from three sensor data types (accelerometry, gyroscope and magnetometer) to classify 12 different natural activities. Three different model architectures (ANN, CNN, Bi-LSTM) are used in the study. FedStack architecture achieves personalization by aggregating heterogenous architectural models at the client level and demonstrates that the FedStack approach consistently outperforms both local and global models.

A recent work in personalized federated learning focused on in-hospital mortality prediction [62]. The study utilized a publicly available electronic health records (HER) database, comprising over 200,000 patients across 208 hospitals in the US. Features were extracted from patients' HER and employed for a binary classification problem, with a multi-layer perceptron (MLP) adopted for modeling. The proposed POLA method involves the initial training of a global federated learning model, referred to as the teacher model. In the subsequent step, local adaptation is accomplished through a Genetic Algorithm (GA) approach. Comparative results highlight the superior performance of the POLA approach against the traditional FedAvg [84] and two other state-of-the-art personalized federated learning architectures [85, 86].

## 3 METHODS AND MATERIALS

This section describes the datasets used in this study, data processing steps and the experimental setup. Further, we describe the prediction model and the different frameworks used in this study.

### 3.1 Clinical datasets

There are two clinical datasets used in this study:

*3.1.1 Ohio T1DM*

The Ohio T1DM [87] dataset was publicly released in two batches (2016 and 2018) with a total of 12 participants. During the 8-week study period, all 12 participants wore a Medtronic Enlite CGM for collecting glucose readings, a Medtronic 530G/ 630G insulin pump and an Empatica/ Basis sensor for collecting physiological data collection. For our analysis, we only focus on the CGM data. The patient data was collected in free-living conditions.

*3.1.2 TCH study*

This is a proprietary data collected at Texas Children's Hospital, Houston, TX. from a total of 113 T1D patients using Dexcom CGM devices. The data for each patient spans a period of 30-90 days. Additional information about this dataset is available in our previous publications. [25].Similar to the OhioT1DM dataset, the data for all patients was collected under free-living conditions.

Comprehensive details about these two datasets are provided in Table I. Figure. 1 illustrates glycemic excursions across all patients, specifically hypoglycemia and hyperglycemia profiles. The dotted grey lines on the x-axis and y-axis represent the median hypoglycemia and hyperglycemia percentages respectively. It is evident that the prevalence of hyperglycemia ($\bar{x} = 41.7\%\ and\ x \cong 22.7\%$) is considerably higher compared to hypoglycemia ($\bar{x} = 2.3\%\ and\ x \cong 1.7\%$).

### 3.2 HH loss

The genesis of our custom loss function is based on two needs: (a) enhancing penalties for errors in glycemic excursion regions, and (b) simultaneously balancing these penalties considering the uneven distribution of samples in the hypoglycemia, normal and hyperglycemia glucose ranges. For the first goal, we add a polynomial increasing penalty for glucose readings further they deviate from the normal range. To achieve the second objective i.e., to account for the disproportionate sample distribution in the hypoglycemia, hyperglycemia, and normal glucose ranges, we introduce a tuning parameter $\alpha$. This parameter ensures that while reducing the errors in the glycemic excursion regions, the predictive performance for overall glucose readings is not compromised.

$$HH\ Loss = \begin{cases} SE, & Y_i \geq 70 \text{ and } Y_i \leq 180 \\ SE + \alpha * penalty, & Y_i < 70 \\ SE + (1-\alpha) * penalty, & Y_i > 180 \end{cases} \quad (1)$$

$$Squared\ Error\ (SE) = (Y_i - \hat{Y}_i)^2,$$
$$penalty = |Y_i - \hat{Y}_i| * (Y_i - c)^2,$$
$$\alpha \in (0,1)$$
$$c = 125\ (midpoint\ of\ glycemic\ excursion\ boundaries)$$

As outlined in Eq. (1), a penalty is applied, equivalent to the square of the distance of the glucose value and the mid-point of the glycemic excursion boundary, for values falling within the hypoglycemia and hyperglycemic ranges.

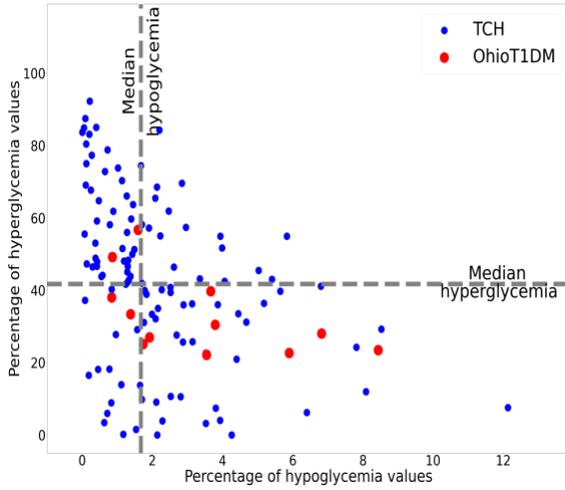

Figure. 1. Glucose profile for patients in TCH study data and OhioT1DM dataset

Table 1: Datasets used in the study

| Dataset | TCH study data | Ohio T1DM |
|---|---|---|
| # of patients | 113 | 12 |
| Median readings per patient | 18873 | 12092 |
| % of hypoglycemic values | 2.8 | 3.19 |
| % of hyperglycemic values | 38.5 | 32.8 |
| CGM Devices Used | Dexcom G6 | Medtronic Enlite |
| Frequency of CGM readings | 5-minutes | 5-minutes |

### 3.3 Federated learning

Federated Learning is a distributed learning framework that enables the training of machine learning models without transmitting sensitive user data to a central server. It relies on collaborative learning between participating users, where each user trains a shared model locally on its own training data through multiple rounds of optimization. Only the model characteristics such as model parameters, weights, gradients etc. are shared among users in the network. The sharing process can occur directly among the participating users or through a central server depending on the FL network topology. After receiving and aggregating the shared model characteristics from all users (e.g.: simple average, weighted averaging), the aggregated information is relayed back to users for further optimization by participants on their individual local data. This iterative communication process continues through multiple rounds until a stopping criterion (e.g.: convergence) is met. All user data remains stored locally, and only specific model characteristics are shared during this collaborative learning process. This approach reduces network and communication costs as model parameters are much smaller in size than actual training data. Additionally, it provides increased accessibility and comparable accuracy to conventional machine learning models. In addition to this, some level of privacy guarantees is ensured. The decentralized nature of federated learning allows patients and clinicians to benefit from more accurate and reliable models trained across a more extensive and diverse data pool.

A general formulation of the federated learning can be expressed as follows:

$$\min_{\phi} L(X_i; \phi),$$

$$L = global\ loss\ function$$

$$K = local\ losses\ \{L_k\}_{k=1}^{K}$$

$$X_k = private\ data$$

### 3.4 Problem formulation

In general, input for a data-driven algorithm for predicting future glucose levels primarily consists of historical glucose readings ($g$) and other features (physiological data, insulin, and carbohydrates intake etc.) if available. In our case where we only rely on glucose readings previously observed and the input data $X_t$ is denoted as:

$$X_t = [x_{t-WL+1}, x_{t-WL+1}, \ldots, x_t] \epsilon\ \mathbb{R}^{1 \times WL}$$

where $x_t \epsilon\ \mathbb{R}^{1 \times 1}$ is the glucose reading at timestep $t$, and $WL$ is the length of previous glucose readings that are considered as input features for prediction. The final set of input features $X$ passed to the model is a transformation $f$ of $X_t$ such that $X = f_N(X_t)$ where $f_N$ is the min-max normalization function to scale the range of input features between the range [0,1].

For a given prediction horizon (PH) and a series of glucose readings at time $t$ as $g_t$, predicting a future glucose value can be defined as

$$\hat{g}_{t+PH} = f_N^{-1}(M(X))$$

where $\hat{g}_{t+PH}$ is an output of the deep learning model $M$ and is de-normalized to get the final prediction.

Based on the available literature [88], data preprocessing involves two main steps: first, replacing 'Low' and 'High' glucose readings with 40 mg/dL and 400 mg/dL, respectively. Second, imputing missing glucose readings through linear interpolation when less than six consecutive readings are missing. This threshold is chosen based on the literature available on autocorrelation between glucose readings [27, 89]. After interpolating missing values, consecutive sequences of glucose readings in the last 2 hours (24 readings) are taken as individual samples. This time-window is selected based on our previous works for glucose prediction [12]. Any samples with missing values at this stage are excluded from our analysis.

For our analysis, we compare three different model types (Figure. 2), frameworks based on the data used for training and the training process:

| **Algorithm 1:** The learning procedure of FedGlu | **Algorithm 2:** Model update for each patient |
|---|---|
| **Input:** Dataset from $N$ distributed patients (clients) $P_1, P_2, ..., P_N$ $\{P_1, P_2, ..., P_N\}$, local minibatch size $B$, number of local epochs $E$, learning rate $\eta$, mean squared error loss $l_{MSE}$, proposed HH loss function $l_{HH}$ | 1 **ClientUpdate**($k, w$): //Run on each client $k$<br>2 $B \leftarrow$ (split $P_k$ into batches of size $B$)<br>3 **for** each local epoch $i$ from 1 to $E$ **do**<br>4   **for** batch $b \in B$ **do**<br>5     $w \leftarrow w - \eta \nabla l_{MSE}(w; b)$<br>6   **end for**<br>7 **end for**<br>8 return the individual trained patient model weights $w$ to server for aggregation |
| 1 /* federated learning process */<br>2 **Cloud server executes:**<br>3 Construct a global federated model $M_G(w)$ with initial weight matrix $W_0$<br>4 **for** each round t = 0,1,2, ... **do**<br>5   Distribute $M_G(w_t)$ to all clients<br>6   **for** each client $k$ **in parallel do**<br>7     $w_{t+1}^k \leftarrow$ **ClientUpdate**($k, w_t$)<br>8   **end for**<br>9   upload all client models to the server<br>10   /* update global model by averaging client models */<br>11   $w_{t+1}^k \leftarrow \sum_{k=1}^{K} \frac{n_k}{n} w_{t+1}^k$<br>12 **end for**<br>13 get the learned global model $M_G(w)$ with weight matrix $W_G$<br>14 distribute the learned model $M_G(w)$ to all clients<br>15 **for** each client $k = 1,2, ... N$ **do**<br>16   $M_G(w_L) \leftarrow$ **LocalFineTune**($D_k, w_S$)<br>17 **end for**<br>**Output:** Locally fine-tuned user model for each client $M_{FT}^k(w_{FT}^k), k \in \{1,2, ..., N\}$ | **Algorithm 3:** Local fine tuning for individual patient<br>1 **LocalFinetune**($k, w$): // For each client k in parallel do<br>2 $B \leftarrow$ (split $P_k$ into batches of size $B$)<br>3 **for** each local epoch $i$ from 1 to $E$ **do**<br>4   **for** batch $b \in B$ **do**<br>5     $w \leftarrow w - \eta \nabla l_{HH}(w; b)$<br>6   **end for**<br>7 **end for**<br>8 return the locally fine-tuned model $M_{FT}(w_{FT})$ with parameter matrix $W_{FT}$ |

*3.4.1 Local model (LM or $M_{Local}$)*

Data is stored locally with the patient and not shared across other entities, such as a central server or other patients in the network. The model is trained individually for each patient, and the training data is limited to individual patient. Local models are fully privacy-preserving with no data or parameter sharing, but they risk lower prediction performance due to limited training data.

*3.4.2 Central model (CM or $M_{Central}$)*

Patient data resides at the central server, and model training occurs on this server. A single model is trained commonly across all available patients. While the central model benefits from a large data pool of data for training, it has minimal privacy preservation for individual patients, as the entire patient data corpus is shared with a central server.

*3.4.3 Federated model (FM or $M_{Fed}$)*

Data storage and model training occur locally at the patient's device. During training, model weights/parameters are frequently shared with a central server for aggregation. These aggregated weights are then returned to individual patients and serve as initializers for further training. Actual data is never shared with other patients in the network or with a global server, maintaining privacy. This is known as model development in a federated-learning framework [84]. Models in this framework achieve better prediction performance through shared learning with other patients while preserving privacy. However, global models may not always be the most suitable for every entity (here, patient) in the network. To address this, we extend the global model via a fine-tuning step, by further optimizing the global model with the custom HH loss function to achieve personalization for individual patients in the network.

### 3.5 FedGlu: The algorithm

The holistic mechanism of the FedGlu algorithm is presented in Algorithm 1 where each patient trains a globally shared model for predicting future glucose values under the coordination of an aggregating server by leveraging federated learning. In each communication round, each client (here, patient) receives a parameterized model from the server, which it optimized using local data (Algorithm 2). After a fixed number of epochs, these individual locally trained model parameters are shared with the server for aggregation. The server aggregates these parameters and shares the updated global model with each individual client. This process repeats until the global model converges. After that, each patient performs an asynchronous 'local fine-tuning' step (Algorithm 3) on the local user data. For this, each client (here, patient) in the networking receives the final global model, which the local clients optimize based on its local data. The loss function is the key difference between training the global model and the local fine-tuning step. The global federated model is trained using the standard MSE loss whereas the local fine-tuning step is performed using the custom HH-loss function.

### 3.6 Model architecture

A multilayer perceptron model is used for analysis in this study. The input layer consists of glucose readings observed in the last 2 hours. This input with dimensions $24 \times 1$ is fed into a dense layer of 512 neurons with a rectified linear unit (ReLU) activation function. The output feature map is passed through two hidden layers with 256 neurons and outputs a 64-neuron layer which, both activated with a ReLU activation. The final output layers get their input from the 3$^{rd}$ dense layer in the network and provide a single number prediction of the future glucose reading.

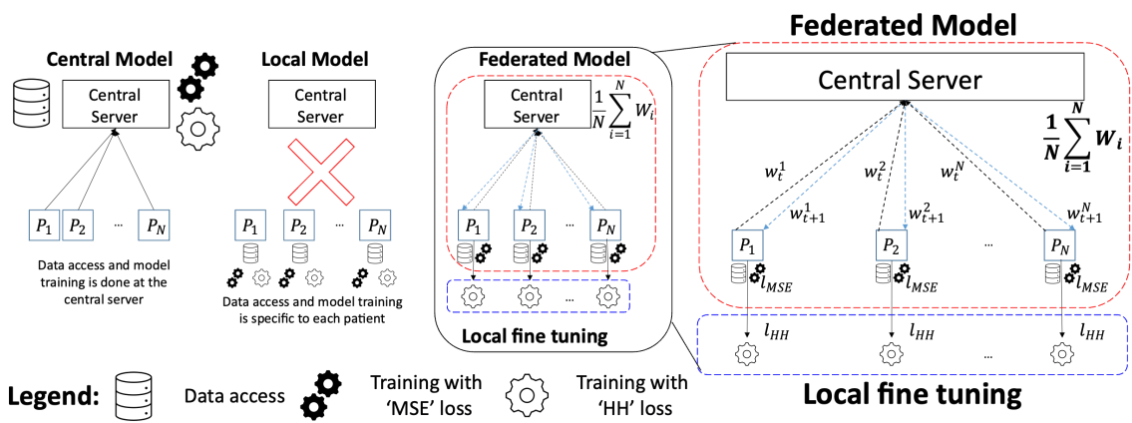

Figure. 2. Visual representation of the different model types/frameworks in this work

### 3.7 Validation approach

In line with our prior research [12, 14], we implement a 5-fold validation strategy with temporally partitioned splits. This strategy guarantees that training and testing splits are derived from non-overlapping time-windows, mitigating potential biases from temporal correlations that could lead to overly optimistic results. This methodology aligns with the BGLP Challenge, where the initial few days of data are utilized for training and the subsequent days constitute the hold-out test set. In our work, we extend the same across multiple splits to make it more robust.

### 3.8 Evaluation metrics

We illustrate the efficiency of our approach using standard metrics in glucose prediction: root mean squared error (RMSE) along with Clark's Error Grid (Figure. 3) [90] to evaluate clinical significance of the predictions. RMSE is defined as:

$$RMSE = \sqrt{\frac{\sum_{i=1}^{N}(y_i - \hat{y}_i)^2}{N}},$$

$N = number\ of\ data\ points$
$y_i = true\ value$
$\hat{y}_i = predicted\ value$

Clarke's Error Grid Analysis (CEGA) is described in Figure. 3 and the definitions of each zone are provided in Table 2.

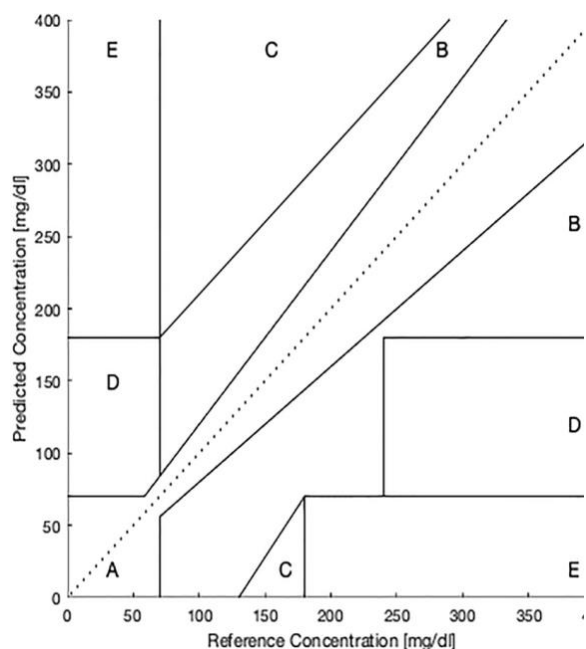

Figure. 3. Reference Clark's Error Grid

Table 2: Clarke's Error Grid Analysis (CEGA) zone-wise description

| Zone | Description |
|---|---|
| A | Predictions within 20% of true values |
| B | Predictions outside of 20% error margin but does not lead to inappropriate treatment |
| C | Predictions leading to unnecessary treatment |
| D | Predictions indicating a potentially dangerous failure to detect hypoglycemia or hyperglycemia |
| E | Predictions that would confuse treatment of hypoglycemia for hyperglycemia and vice versa |

### 3.9 Model training/ Hyperparameter tuning

To provide robust estimates of our methodology and support the enlisted contributions with results, we consider the following experiments:

1. Setup 1 – Advantage of HH loss function: The performance with the HH loss function is compared against the standard mean squared error (MSE) loss. We demonstrate the advantage in terms of standard regression metrics and the clinical significance for glucose predictions. We compare the HH loss function results across two separate datasets namely TCH study data and the OhioT1DM dataset. We use a central (population-level) model to obtain the results.

2. Setup 2 – Usefulness of Personalized Federated Models: We compare the performance across the different model types: central model, local model, global federated model, and the personalized federated model. The comparison is made independently across both datasets. This setup will prove the robustness of ML models trained in a federated learning work against the central and local models.

### 3.10 Evaluation metrics

All models at central, local, and federated (global and personal) levels shared identical architecture and parameters. The learning rate was fixed at 0.001 with a constant batch-size of 500. The training was set to a maximum of 50 epochs, with early stopping for a patience of 10 epochs to avoid prevent redundant training and thereby reduce computational time. The global federated model is trained with TensorFlow-federated. The client-optimizer was set to 'Adam' with a learning rate of 0.001 and the serve-optimizer to 'SGD' with a learning rate of 1 to mimic the baseline federated model [84]. The number of communication rounds was set to 50, and a simple weighted-average proportional to the number of samples with each patient (node) in the network was used as the aggregating function. The global federated model is saved after observing the train loss (MSE) convergence. Individual patients (nodes) then fine-tune this saved global federated model on their individual local data with an 'Adam' optimizer and a learning rate of 0.001. This fine-tuning step is however done the custom HH loss function. For comparing results, the local-level and (personalized) federated-level models, with a specific $\alpha$, is selected based on the training data where the combined RMSE (across hypoglycemia and hyperglycemia) is minimum. At the central level, a single $\alpha$ which showed the maximum combined reduction across all patients, was considered.

## 4 RESULTS

This section describes the datasets used in this study, data processing steps and the experimental setup. Further, we describe the prediction model and the different frameworks used in this study.

### 4.1 Advantage of HH loss function

In the first analysis, we assess how the HH loss functions enhances prediction performance in the glycemic excursion regions while maintaining clinical significance for overall predictions based on two independent datasets. For this we consider a centralized (population-level) model development setting. Since a common model is trained across all the patients for the dataset, we choose a single $'\alpha'$ value ($\alpha = 1$) which showed the combined maximum improvement (reduction in RMSE) across hypoglycemia and hyperglycemia ranges on the training dataset.

*4.1.1 TCH study dataset*

Figure. 4 (left), gives a performance comparison between the HH loss and the baseline mean squared error (MSE) loss. The HH loss function exhibits a 52% reduction in root-mean-squared error (RMSE) for the hypoglycemia region compared to MSE, while showing similar performance in the hyperglycemia region. Although there is a slight dip in overall prediction performance (regarding RMSE values), the clinical impact evaluated through Clark's EGA is negligible and on the contrary better with HH loss than MSE.

In our evaluation using Clark's Error Grid (Table 3) to gauge the clinical relevance of the predictions, the HH loss exhibited superior performance, missing to detect an average of only 0.51% of excursions, compared to 2.07% with MSE - representing a substantial 75% reduction. This refers to Zone: D+E from CEGA, and predictions falling in this region are analogous to false negatives from the confusion matrix used to detect hypoglycemia and hyperglycemia. Excursion detection is critically important as it allows patients to take intervention measures based on the prediction. Additionally, overall glucose predictions within the combined Region A+B improved from 97.94% to 99.39%. These findings underscore that, while the HH loss introduces a bias (increasing predictions in Zone: C) towards enhancing accuracy in excursion regions (hypoglycemia and hyperglycemia), it does not detrimentally affect the clinical accuracy of overall glucose predictions.

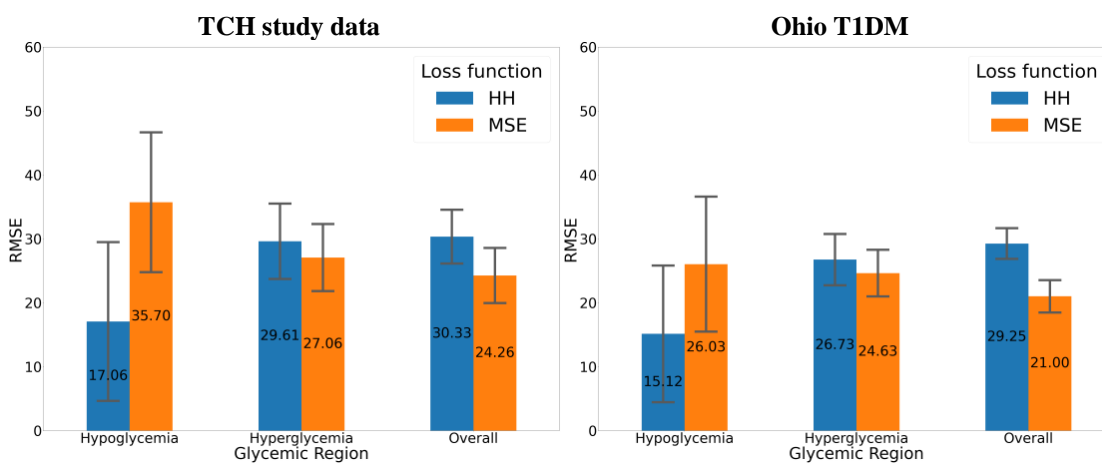

Figure. 4. Performance comparison: MSE vs HH vs NMSE for (left) TCH study data and (right) OhioT1DM dataset

Table 3: Summary of Clarke's Error Grid Analysis (CEGA) performance for different loss functions

| Loss | TCH study data | | | Ohio T1DM | | |
|---|---|---|---|---|---|---|
| | Zone: A+B (TP & TN) | Zone: C (FP) | Zone: D + E (FN) | Zone: A+B (TP & TN) | Zone: C (FP) | Zone: D + E (FN) |
| MSE | 97.84±8.78 | 0.09±0.11 | 2.07±1.68 | 97.89±7.29 | 0.05±0.04 | 2.06±1.30 |
| HH | 99.36±0.14 | 0.13±0.14 | 0.51±0.32 | 99.56±8.25 | 0.06±0.05 | 0.38±0.19 |
| Difference (in %) | 1.56↑ | 44.44↑ | 75.36↑ | 1.71↑ | 20.00↑ | 81.56↑ |

### 4.1.2 Ohio T1DM dataset

For the OhioT1DM dataset, Figure. 4 (right) a similar performance comparison, the HH loss exhibits an RMSE of 15.12 for the hypoglycemia region, which is 41% less than the MSE, whereas in the hyperglycemia region the RMSE remains relatively consistent. Similar to the performance observed with the TCH study data, there is a decline in overall prediction performance in terms of RMSE values. However, this performance drop is trivial when assessed for the clinical significance with CEGA. The CEGA provides evidence for this claim. The number of data points in the regions: D+E from reduces from 2.05% with MSE to 0.32% (84% reduction) with the HH loss – indicating an increase in the model's ability to predict excursions. Concurrently, the percentage of data points in the regions: A+B increases from 97.89% to 99.59%, suggesting an improvement in overall predictions without compromise.

**4.2 Model training in a federated learning framework**

The predictive capabilities of models incorporating the HH loss function were extensively evaluated at the local, central, and federated levels. In Figure. 5, the magnitude of difference (measured in RMSE) in various glycemic excursion regions is presented. Table 4 compares different model types through Clarke's Error grid analysis, while Table 5 outlines improvements in terms of the number of patients for RMSE and CEGA. Compared to local models, federated models exhibit an improvement of 16.67% (in RMSE) for predicting hypoglycemia and 18.91% (in RMSE) for predicting hyperglycemia simultaneous – both statistically significant using the paired t-test ($p \ll 0.01$). For the TCH study data, federated models demonstrated improvements for 96 (out of 113) patients in hypoglycemia region and 110 (out of 111) patients in hyperglycemia regions over local models simultaneously. Regarding the OhioT1DM dataset, federated models improved by 9% (in RMSE) for predicting hypoglycemia and 33.29% (in RMSE) for predicting hyperglycemia. Although the improvement in hyperglycemia is statistically significant ($p \ll 0.01$) but hypoglycemia ($p = 0.18$) is not. For the OhioT1DM dataset, federated models improve for 9 (out of 12) patients in hypoglycemia region and 12/12 patients in hyperglycemia region, over the local models. Across the two datasets, there is an improvement of 12.37% (in RMSE) for predicting hypoglycemia and 29.05% (in RMSE) for prediction hyperglycemia – which are both statistically significant with ($p \ll 0.01$).

When evaluated for clinical significance through Clark's Error Grid, it is evident that HH loss function improves glycemic excursions (Region: D+E) prediction at all levels without compromising the clinical significance of overall predictions (Region: A+B) across both datasets. Comparing local and federated models, for the TCH study data, we see a 37% reduction *($p \ll 0.01$)* and for the OhioT1DM dataset, 31% reduction *($p \ll 0.01$)* in points falling in the Region: D+E (compared to MSE) which signifies an increase in the detection capability of glycemic excursions. When comparing local and federated models for predictions falling in the Region: C of CEGA, for the TCH study data, there is a 50% reduction whereas for the OhioT1DM dataset, there is a 20% reduction in which signifies fewer false predictions with the federated models. Table 4 also provides evidence that federated models when compared to local models through CEGA, are able to improve over local models for all regions of CEGA for almost all patients across the two datasets.

On the other hand, central models distinctly outperform federated models for both hypoglycemia and hyperglycemia due to their advantage of a significantly larger training pool. This is evident in the evaluation with RMSE and CEGA. However, compared to the local models, federated models achieve a much closer performance than central models. Table 5 outlines the improvement with federated models against central and local models for each of TCH and OhioT1DM datasets.

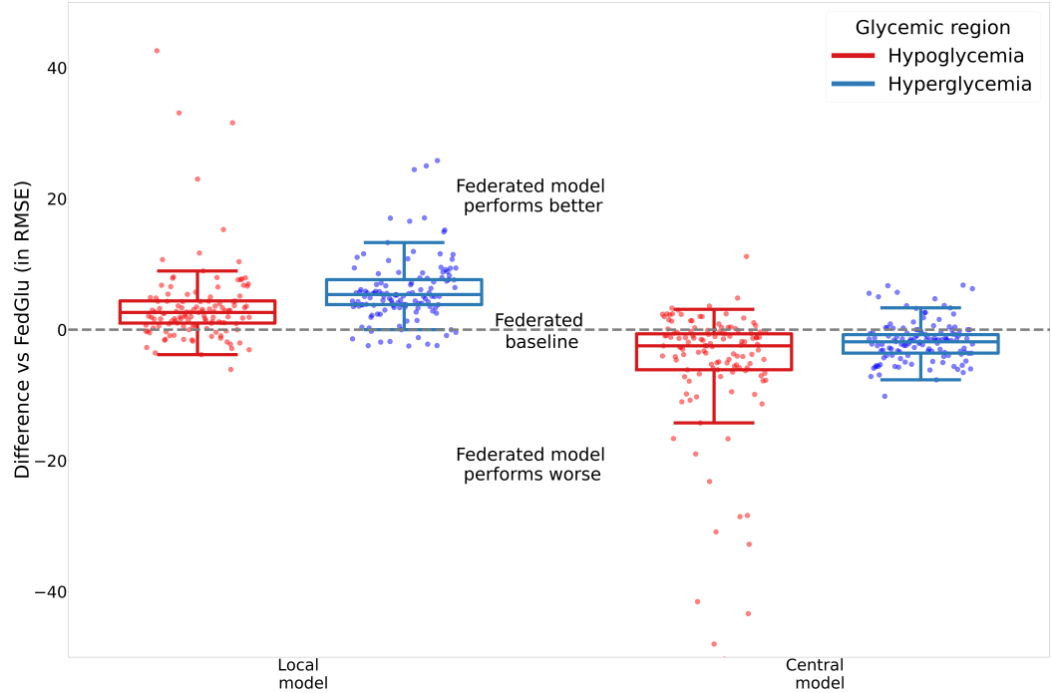

Figure. 5. Comparing improvements with federated Model against local and central models for hypoglycemia and hyperglycemia

Table 4: Summary of Clarke's Error Grid Analysis (CEGA) performance for different model types

| Model | TCH study data | | | Ohio T1DM | | |
|---|---|---|---|---|---|---|
| | Zone: A+B (TP & TN) | Zone: C (FP) | Zone: D + E (FN) | Zone: A+B (TP & TN) | Zone: C (FP) | Zone: D + E (FN) |
| Central | 99.36±13.22 | 0.13±0.14 | 0.51±0.32 | 99.56±8.25 | 0.06±0.05 | 0.38±0.19 |
| Local | 98.51±19.12 | 0.31±0.33 | 1.18±0.77 | 99.19±11.09 | 0.10±0.06 | 0.72±0.25 |
| Federated (FedGlu) | 99.12±16.15 | 0.15±0.15 | 0.74±0.40 | 99.42±7.34 | 0.08±0.06 | 0.50±0.32 |

Table 5: Federated model improvements over (a) local and (b) central models

| Comparing for RMSE | | | | |
|---|---|---|---|---|
| FedGlu Improvement over | Condition | Dataset | | |
| | | TCH | OhioT1DM | Combined |
| Local model | Hypoglycemia | 96/113 | 9/12 | 105/125 |
| | Hyperglycemia | 110/111 | 12/12 | 122/123 |
| Central model | Hypoglycemia | 19/113 | 2/12 | 21/125 |
| | Hyperglycemia | 13/111 | 2/12 | 15/123 |
| Comparing for excursion detection (CEGA) | | | | |
| FedGlu Improvement over | CEGA Region | Dataset | | |
| | | TCH | OhioT1DM | Combined |
| Local model | A+B | 109/113 | 12/12 | 121/125 |
| | D+E | 109/113 | 12/12 | 121/125 |
| | C | 108/113 | 9/12 | 117/125 |
| Central model | A+B | 13/113 | 0/12 | 13/125 |
| | D+E | 13/113 | 0/12 | 13/125 |
| | C | 58/113 | 6/12 | 64/125 |

## 5 DISCUSSION

Based on the results, the main contributions of our work can be summarized as:

1. Introduction of a novel HH loss function aimed at improving glycemic predictions in the excursion regions while ensuring clinically significance.
2. Model implementation with the HH loss function within a federated learning framework, balancing performance in the excursion region and privacy.

### 5.1 Insights and observations

Models with HH loss function can predict more accurately and clinically significantly in glycemic excursion regions at local, central, and federated levels. This performance may result in sub-optimal RMSE values in the overall glucose ranges. However, when compared for clinical significance, there is no reduction and on the contrary, it improves overall predictions in regions: A+B. A detailed evaluation with CEGA with different model types, including local, central, and federated models is presented in Table 6. A further expansion of CEGA zones for Table 6 is provided in Appendix 1.

In the proposed HH loss function, the parameter '$\alpha$' plays a vital role in penalizing the errors and achieving balance across the three different glycemic regions. This can be optimally chosen for the local and federated models as model training is done locally. However, in the case of the central model, we can choose a single value for the entire cohort of patients. This may be beneficial to the majority of patients but not all the patients. We further explore the impact of choosing different '$\alpha$' values for the central model.

Also, when we compare model performance across the different model types, we see that federated model has varying level of differences compared to local and central models. We try to investigate further these improvements concerning the glycemic profiles of patients and who is likely to benefit the most.

Table 6: Summary of Clarke's Error Grid across different (a) model types and (b) loss functions

| Model | Loss | TCH | | |
|---|---|---|---|---|
| | | Zone: A+B (TP & TN) | Zone: C (FP) | Zone: D + E (FN) |
| Central | MSE | 97.84±8.78 | 0.09±0.11 | 2.07±1.68 |
| | HH | 99.36±13.22 ↑ | 0.13±0.14 ↓ | 0.51±0.32 ↓ |
| Local | MSE | 97.35±11.23 | 0.26±0.37 | 2.39±1.71 |
| | HH | 98.51±19.12 ↑ | 0.31±0.33 ↓ | 1.18±0.77 ↓ |
| Federated | MSE | 97.37±9.49 | 0.10±0.11 | 2.53±2.02 |
| | HH | 99.12±16.15 ↑ | 0.15±0.15 ↓ | 0.74±0.40 ↓ |
| Model | Loss | OhioT1DM | | |
| | | Zone: A+B (TP & TN) | Zone: C (FP) | Zone: D + E (FN) |
| Central | MSE | 97.89±7.29 | 0.05±0.04 | 2.06±1.30 |
| | HH | 99.56±8.25 ↑ | 0.06±0.05 ↓ | 0.38±0.19 ↓ |
| Local | MSE | 97.59±7.87 | 0.06±0.05 | 2.33±1.32 |
| | HH | 99.19±11.09 ↑ | 0.10±0.06 ↓ | 0.72±0.25 ↓ |
| Federated | MSE | 97.60±7.83 | 0.05±0.04 | 2.36±1.61 |
| | HH | 99.42±7.34 ↑ | 0.08±0.06 ↓ | 0.50±0.32 ↓ |

## 5.2 Impact of '$\alpha$' parameter

The parameter '$\alpha$' can be tuned to achieve a well-balanced optimal prediction performance for hypoglycemia and hyperglycemia excursions regions simultaneously. A higher '$\alpha$' value prioritizes performance in the hypoglycemia regions whereas a lower '$\alpha$' emphasizes performance in the hyperglycemia region. Figure. 6 illustrates the average improvement (in RMSE) across patients in both glycemic excursion regions occurring co-occurring for various '$\alpha$' values. The red curve, representing improvement in hypoglycemia values, shows an exponential increase compared to MSE as '$\alpha$' increases. In contrast, the blue curve indicating improvement in hyperglycemia values, exhibits a downward trend with increasing '$\alpha$'. However, the slope for hypoglycemia values is significantly steeper than hyperglycemia values. This is because of the high data imbalance that exists, especially for hypoglycemia values (median: 1.64%) as compared to hyperglycemia values (median: 43%). This also highlights the greater of accurately predicting hypoglycemia values compared to hyperglycemic values. The parameter '$\alpha$' can be customized for local and federated models based on individual preferences and glycemic profiles to yield optimal results. Table 7 provides performance metrics (for central models) in terms of Clark's EGA for different '$\alpha$' values.

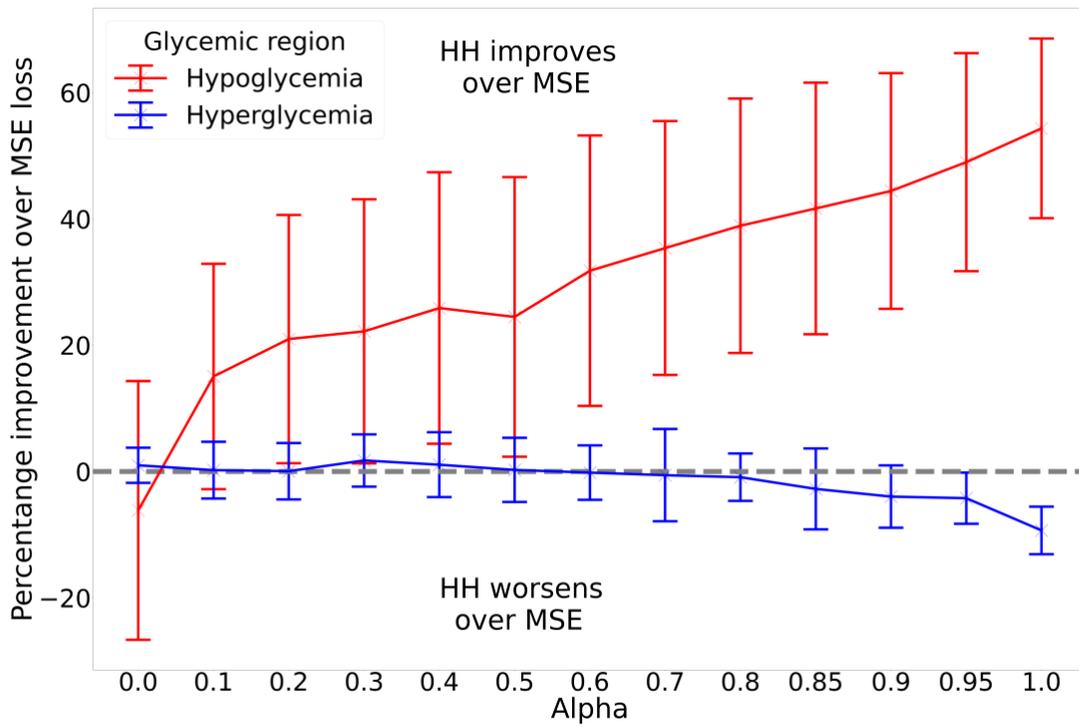

Figure. 6. Improvement in glycemic excursion regions with HH loss over MSE for the Central Model

Table 7: Central model performance with different values of '$\alpha$' parameter (TCH study + Ohio T1DM data)

| Alpha | Clark's EGA Analysis | | # of patients HH loss improves over MSE for | |
|---|---|---|---|---|
| | Zone: A+B (True predictions) | Zone: C+D+E (False predictions) | Hypoglycemia (125 patients) | Hyperglycemia (123 patients) |
| 0 | 97.58 | 0.41 | 64 | 93 |
| 0.1 | 98.92 | 1.08 | 101 | 70 |
| 0.2 | 99.11 | 0.89 | 108 | 60 |
| 0.3 | 99.18 | 0.82 | 107 | 99 |
| 0.4 | 99.19 | 0.81 | 111 | 90 |
| 0.5 | 99.21 | 0.79 | 109 | 78 |
| 0.6 | 99.28 | 0.72 | 115 | 67 |
| 0.7 | 99.33 | 0.68 | 117 | 62 |
| 0.8 | 99.33 | 0.67 | 120 | 53 |
| 0.85 | 99.34 | 0.66 | 121 | 27 |
| 0.9 | 99.36 | 0.65 | 124 | 18 |
| 0.95 | 99.39 | 0.61 | 124 | 14 |
| 1 | 99.37 | 0.63 | 125 | 0 |
| MSE | 97.78 | 2.21 | - | - |

Table 8: Central model performance with different values of '$\alpha$' parameter (TCH study + Ohio T1DM data)

| Loss | TCH study data | | | Ohio T1DM | | |
|---|---|---|---|---|---|---|
| | Zone: A+B (TP & TN) | Zone: C (FP) | Zone: D + E (FN) | Zone: A+B (TP & TN) | Zone: C (FP) | Zone: D + E (FN) |
| MSE | 97.84±8.78 | 0.09±0.11 | 2.07±1.68 | 97.89±7.29 | 0.05±0.04 | 2.06±1.30 |
| NMSE | 98.47±8.44 | 0.06±0.08 | 1.47±0.96 | 98.87±6.84 | 0.02±0.02 | 1.10±0.58 |
| HH | 99.36±13.22 ↑ | 0.13±0.14 ↓ | 0.51±0.32 ↓ | 99.56±8.25 ↑ | 0.06±0.05 ↓ | 0.38±0.19 ↓ |

Table 9: Central model performance with different values of '$\alpha$' parameter (TCH study + Ohio T1DM data)

| Paper | Year | A+B | C+D+E |
|---|---|---|---|
| Using grammatical evolution to generate short-term blood glucose prediction models [91] | 2018 | 98.60 | 1.40 |
| Blood glucose prediction for type 1 diabetes using generative adversarial networks [92] | 2020 | 99.54 | 2.69 |
| The Diabits app for smartphone-assisted predictive monitoring of glycemia in patients with diabetes: retrospective observational study [93] | 2020 | 99.07 | 0.93 |
| Comparing machine learning techniques for blood glucose forecasting using free-living and patient generated data [94] | 2020 | 99.36 | 0.64 |
| Adversarial multi-source transfer learning in healthcare: Application to glucose prediction for diabetic people [95] | 2021 | 99.20 | 0.80 |
| GLYFE: review and benchmark of personalized glucose predictive models in type 1 diabetes [96] | 2022 | 89.73 | 10.27 |
| Deep representation-based transfer learning for deep neural networks [97] | 2022 | 98.34 | 1.66 |
| An evolution-based machine learning approach for inducing glucose prediction models [98] | 2022 | 98.43 | 1.57 |
| A multitask learning approach to personalized blood glucose prediction [47] | 2022 | 99.10 | 0.90 |
| Personalized blood glucose prediction for type 1 diabetes using evidential deep learning and meta-learning [99] | 2023 | 98.86 | 1.24 |
| Glucose Transformer: Forecasting Glucose Level and Events of Hypoglycemia and Hyperglycemia [46] | 2023 | 99.89 | 0.12 |
| Short-term prediction method of blood glucose based on temporal multi-head attention mechanism for diabetic patients [100] | 2023 | 99.63 | 0.36 |
| Deep multitask learning by stacked long short-term memory for predicting personalized blood glucose concentration [45] | 2023 | 99.41 | 0.61 |
| Forecasting with sparse but informative variables: A case study in Predicting blood glucose [101] | 2023 | 99.23 | 0.77 |
| ***Our Approach (Central model)*** | *-* | ***99.56*** | ***0.44*** |
| ***Our Approach (FedGlu)*** | *-* | ***99.42*** | ***0.58*** |

### 5.3 Federated models: Who benefits the most?

We compare the performance improvements achieved with federated models against central and local models (with HH loss) across varying glycemic profiles of both hypoglycemia and hyperglycemic regions. To categorize patients effectively, they are initially binned into groups of ten based on the percentage of hypoglycemic or hyperglycemic values in their profile. For hypoglycemia, these intervals include (0, 0.22]%, (0.22, 0.5]%, and so forth, while for hyperglycemia, intervals are defined as (0, 7.81]%, (7.81, 19.31]%, and so on.

In the context of hypoglycemia prediction (Figure. 7 - left), a notable trend emerges: as the percentage of hypoglycemic values increases, the prediction performance of federated models, global models and local models converges. Patients with higher hypoglycemia values exhibit similar prediction performance across local, central, and federated models. However, for patients with extremely low instances of hypoglycemia, central models far outperform local and federated models. This is because training data for central models has multiple instances of hypoglycemia for the model to train whereas local and federated models do not have that advantage. The Spearman correlation coefficient ($\rho$) between mean improvement (in RMSE) with federated models and increasing hypoglycemic (interval) values, is statistically significant against both local ($\rho = 0.01$) and central models ($\rho = 0.02$). When compared for variance, it shows similar trend and is statistically significant across local ($\rho = 0.01$) and central ($\rho = 0.01$) models. It is also observed that, predictions performance with federated models is closer to the central models and also maintain a clear advantage over the local models for close to 50% of the patients.

In contrast, for hyperglycemia prediction (Figure. 7 - right), no discernible patterns emerge across the different model types (federated model, global model, and local model) when compared to the hyperglycemia profile of patients. The Spearman correlation coefficient ($\rho$) for improvements with federated models over local ($\rho = 0.7$) and central models ($\rho = 0.1$) is not statistically significant. Furthermore, the performance disparity across these three model types (relative to hypoglycemia) is minimal for hyperglycemia prediction. A major contributing factor is a substantial difference in the number of hypoglycemic glucose values compared to hyperglycemic glucose values (the lowest bin for hyperglycemia is (0, 7.81]% and a median of 1.64%. In contrast, the highest bin for hypoglycemia is (4,63, 12.13]% and a median of 43%).

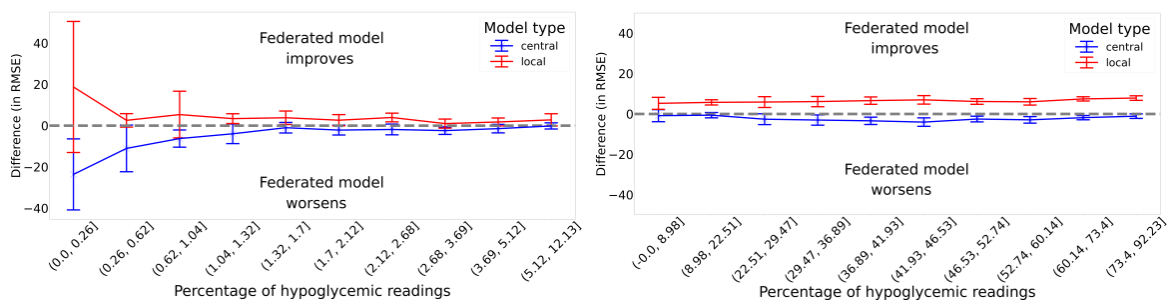

Figure. 7. Federated Model performance vs Local and Central Models across different patient profiles in (left) hypoglycemia region and (right) hyperglycemia region

### 5.4 Comparison with literature

We comprehensively compared between the HH loss function and a recent state-of-the-art NMSE (normalized mean squared error) that showed significant performance improvement in the hypoglycemic range. The comparison is made on standard evaluation metrics of RMSE and Clark's EGA. Analyzing the TCH study data, HH loss improves performance against NMSE (Figure. 8 - left) by 29%. ($p \ll 0.01$) in the hypoglycemia region and by 5% ($p \ll 0.01$) in the hyperglycemia region. Similarly, for the OhioT1DM dataset (Figure. 8 - right), HH loss improves performance (in RMSE) in the hypoglycemia region by 21%. ($p \ll 0.01$) and hyperglycemia region by 6% ($p \ll 0.01$). Moreover, when compared using Clark's EGA, models with the HH loss function fails to detect only 0.51% of glycemic excursions compared to 1.47% with NMSE (a 65% reduction). We also see an increase in points falling in Regions: A+B with the HH loss function compared to using either of MSE or NMSE (Table 8).

The OhioT1DM dataset, used in the BGLP challenge, is a standard dataset for comparing glucose prediction performance. While the BGLP challenge and other literature aim to show accurate predictions of overall glucose values, our work focuses explicitly on improved performance in the glycemic excursion regions (hypoglycemia and hyperglycemia), making comparisons challenging. Nevertheless, we attempt to show our work in the light of other notable studies based on CEGA, providing a standard and fair way to assess clinical significance.

Table 9 compares research achieving state-of-the-art performances using the OhioT1DM dataset. Remarkably, our simple MLP model, with several hundred times lower number of training parameters, outperforms or matches the top-performing models reported. Additionally, federated models, lacking access to the entire training data, show promising results compared to other state-of-the-art methods proposed in the literature.

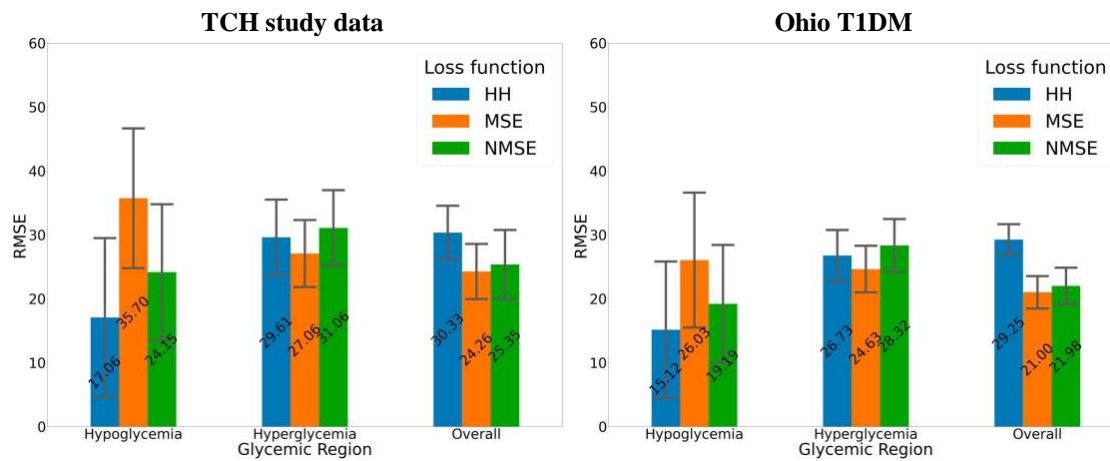

Figure. 8. Performance comparison: MSE vs HH vs NMSE for (left) TCH study data and (right) Ohio T1DM

### 5.5 Advantages and limitations in the proposed approaches

We introduce a bias towards glycemic excursion regions to enhance performance in the hypoglycemia and hyperglycemia regions. As a result, the prediction error for overall glucose does increase. However, we overcome this limitation by ensuring that clinical significance is not compromised. Additionally, when using HH loss, the federated models undergo a two-step construction process: (i) initial training with MSE loss function and (ii) subsequent fine-tuning with the HH loss at local levels. In future endeavors, we aim to implement a single step personalized federated model combining these two phases. This work uses the baseline federated learning framework to develop the proposed algorithm, FedGlu. There can be many challenges for its adoption in the real-world, like the unavailability of one or more entities in the network, a corrupt local entity or a corrupt central server which may compromise the privacy of the framework. However, this work aimed to lay the groundwork for using federated learning in glucose prediction algorithms that can benefit patients. The authors plan to improve the proposed FedGlu by adding privacy guarantees in subsequent works and also tackle the challenge of unavailability among participating entities for improved learning.

## 6 CONCLUSION

In this work, we proposed a novel HH loss function that simultaneously improves predictions in both hypoglycemia and hyperglycemia regions without compromising overall predictions. This is critical as the HH loss function improves the glycemic excursion detection capabilities by an average of 78% across two datasets (a total of 125 patients) as compared to the standard MSE loss. The results were consistent with a proprietary dataset and publicly available OhioT1DM datasets. We also demonstrated the consistency of the HH loss function via FedGlu, a machine learning through a collaborative learning approach. FedGlu clearly outperforms the local models (35% improved glycemic excursion detection capabilities) while coming close to central model performances. These results prove the need to develop machine learnings models with strong predictive capabilities and ensure privacy for sensitive patient healthcare data.

# A APPENDICES

Appendix 1: Central model performance with different values of '$\alpha$' parameter (TCH study + Ohio T1DM data)

| Model | Loss | TCH study data | | | | | |
|---|---|---|---|---|---|---|---|
| | | *C-Lower* | *C-Upper* | *D-Left* | *D-Right* | *E-Right lower* | *E-Left Upper* |
| Central | MSE | 0.000 | 0.085 | 1.858 | 0.212 | 0.000 | 0.003 |
| | HH | 0.024 | 0.108 | 0.151 | 0.237 | 0.114 | 0.003 |
| Local | MSE | 0.001 | 0.255 | 1.962 | 0.413 | 0.003 | 0.013 |
| | HH | 0.029 | 0.276 | 0.483 | 0.606 | 0.081 | 0.012 |
| Federated | MSE | 0.000 | 0.095 | 2.229 | 0.299 | 0.001 | 0.005 |
| | HH | 0.017 | 0.132 | 0.378 | 0.323 | 0.052 | 0.004 |
| **Model** | **Loss** | **OhioT1DM dataset** | | | | | |
| | | *C-Lower* | *C-Upper* | *D-Left* | *D-Right* | *E-Right lower* | *E-Left Upper* |
| Central | MSE | 0.000 | 0.051 | 1.913 | 0.131 | 0.002 | 0.01 |
| | HH | 0.007 | 0.051 | 0.156 | 0.163 | 0.056 | 0.005 |
| Local | MSE | 0 | 0.064 | 2.139 | 0.189 | 0.002 | 0.013 |
| | HH | 0.003 | 0.093 | 0.461 | 0.217 | 0.028 | 0.011 |
| Federated | MSE | 0 | 0.053 | 2.158 | 0.185 | 0.004 | 0.01 |
| | HH | 0.002 | 0.073 | 0.328 | 0.138 | 0.023 | 0.01 |